\documentclass{article}

    \PassOptionsToPackage{numbers, compress}{natbib}
    \usepackage[final]{neurips_2022_ml4ad} 

\usepackage[utf8]{inputenc} 
\usepackage[T1]{fontenc}    
\usepackage{hyperref}       
\usepackage{url}            
\usepackage{booktabs}       
\usepackage{amsfonts}       
\usepackage{nicefrac}       
\usepackage{microtype}      
\usepackage{xcolor}         

\usepackage{enumitem}
\usepackage{float}
\usepackage{etoolbox}
\makeatletter
\patchcmd{\@algocf@start}
  {-1.5em}
  {0pt}
  {}{}
\makeatother
\usepackage{multicol}
\usepackage{svg}

\usepackage{tikz}
\usepackage{arydshln}
\usepackage[symbol]{footmisc}
\usepackage{footnote}
\usepackage{graphicx}
\usepackage{wrapfig,lipsum}
\usepackage{amssymb}
\usepackage{setspace}
\usepackage{comment}

\hypersetup{colorlinks,citecolor={teal}}

\usepackage{caption}
\usepackage{subcaption}
\usepackage{supertabular}
\usepackage{array}
\usepackage{amsmath}
\usepackage{bm}
\usepackage{multirow}

\usepackage{longtable}

\definecolor{Red}{rgb}{0.6,0,0}
\definecolor{Blue}{rgb}{0,0,0.8}
\definecolor{Green}{rgb}{0,0.4,0.7}
\definecolor{airforceblue}{rgb}{0.36, 0.54, 0.66}
\definecolor{ao(english)}{rgb}{0.0, 0.5, 0.0}
\definecolor{azure(colorwheel)}{rgb}{0.0, 0.5, 1.0}
\definecolor{crimson}{rgb}{0.86, 0.08, 0.24}
\definecolor{darkcerulean}{rgb}{0.03, 0.27, 0.49}
\definecolor{cobalt}{rgb}{0.0, 0.28, 0.67}
\definecolor{rosegold}{rgb}{0.72, 0.43, 0.47}
\definecolor{orange-red}{rgb}{1.0, 0.27, 0.0}
\definecolor{mountainmeadow}{rgb}{0.19, 0.73, 0.56}
\definecolor{malachite}{rgb}{0.04, 0.85, 0.32}
\definecolor{darkblue}{rgb}{0.0, 0.0, 0.55}
\definecolor{navyblue}{rgb}{0.0, 0.0, 0.5}
\definecolor{customblue}{rgb}{0.2, 0.35, 0.8}

\hypersetup{colorlinks=true}
\hypersetup{linktoc=all}
\hypersetup{citecolor=darkcerulean}
\hypersetup{linkcolor=crimson}
\hypersetup{urlcolor=black}
\usepackage[all]{hypcap}

\usepackage{placeins}
\usepackage[hyphenbreaks]{breakurl}
\usepackage{xurl}
\makesavenoteenv{tabular}
\makesavenoteenv{table}
\usepackage{amsthm}

\usepackage{algorithm}
\usepackage{algorithmic}

\title{Distortion-Aware Network Pruning and \\ Feature Reuse for Real-time Video Segmentation}

\author{
  Hyunsu Rhee$^{1}$,\hspace{0.05in}
  Dongchan Min$^{1}$,\hspace{0.05in}
  Sunil Hwang$^{1}$,\hspace{0.05in} \\
  \textbf{Bruno Andreis$^{1}$,}\hspace{0.05in}
  \textbf{Sung Ju Hwang$^{1,2}$} \\
  KAIST$^{1}$,
  AITRICS$^{2}$,
  South Korea \\
  \texttt{\{ryanrhee, alsehdcks95, sunilhoho, andries, sjhwang82\}@kaist.ac.kr} \\
}

\begin{document}

\maketitle

\begin{abstract}
Real-time video segmentation is a crucial task for many real-world applications such as autonomous driving and robot control. Since state-of-the-art semantic segmentation models are often too heavy for real-time applications despite their impressive performance, researchers have proposed lightweight architectures with speed-accuracy trade-offs, achieving real-time speed at the expense of reduced accuracy. In this paper, we propose a novel framework to speed up any architecture with skip-connections for real-time vision tasks by exploiting the temporal locality in videos. Specifically, at the arrival of each frame, we transform the features from the previous frame to reuse them at specific spatial bins. We then perform partial computation of the backbone network on the regions of the current frame that captures temporal differences between the current and previous frame. This is done by dynamically dropping out residual blocks using a gating mechanism which decides which blocks to drop based on inter-frame distortion. We validate our Spatial-Temporal Mask Generator (STMG) on video semantic segmentation benchmarks with multiple backbone networks, and show that our method largely speeds up inference with minimal loss of accuracy.
\end{abstract}
\section{Introduction} \label{introduction}
Semantic segmentation~\citep{zhao2017pyramid, zhao2018icnet, yu2018bisenet, chen2019fasterseg, li2020semantic, fan2021rethinking, hong2021deep} is a fundamental task in computer vision that is crucial for many real-world applications including autonomous driving, robot control, augmented reality, surveillance system, aerial imagery, drone image analysis, and medical diagnosis~\citep{ronneberger2015u, cordts2016cityscapes, zhou2018unet++, balloch2018unbiasing, azimi2019skyscapes, oh2020segmenting, jung2021standardized}.
Recently, much focus has been placed on developing segmentation models for streaming videos. This is a challenging task that requires not only high accuracy but also real-time inference speed~\citep{hu2020temporally, wang2021temporal}.  
However, majority of semantic segmentation models still rely on computationally heavy backbone networks~\citep{he2016deep} to achieve state-of-the-art performance. These heavy backbones, in most cases, become computational bottlenecks in the real-time video semantic segmentation pipeline.

\begin{figure*}
    \centering
    \vspace{-0.1in}
        \includegraphics[width=\linewidth]{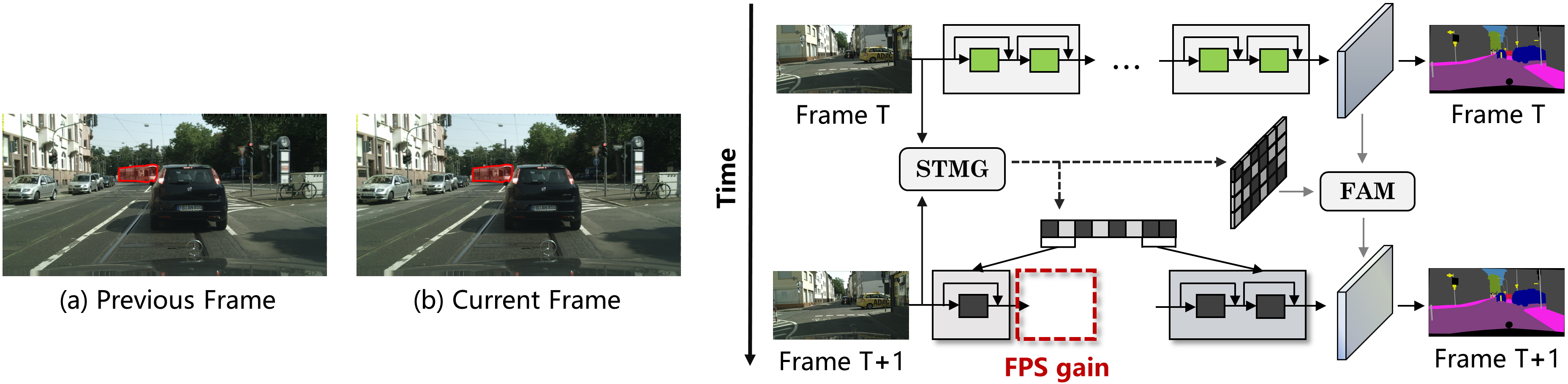}
    \vspace{-0.1in}
    \caption{\textbf{Conceptual Overview.} Our proposed Spatial-Temporal Mask Generator (STMG) captures temporally evolving regions (highlighted by the red box) between two consecutive frames using spatial masks. This allows us to reuse the computed features from the previous frames for static regions. Additionally, conditioned on the distortion between the two frames, we generate a block pruning mask that determines which residual blocks to drop when running the segmentation model on the current frame. The transformed features from the previous frame are then combined with the output of the partial network computation using a feature aggregation module (FAM).}
    \label{fig:1}
    \vspace{-0.2in}
\end{figure*}

One approach to tackle this challenge is to identify the key frames which are processed with heavy backbone networks and to propagate the pre-computed features of key frames for other frames~\citep{zhu2017deep, jain2019accel, wang2020dynamic, zhuang2020video, xu2018dynamic} in the video stream. However, this approach is based on the assumption that flow estimation and feature propagation are faster than running the backbone networks~\citep{zhu2017deep}. Since state-of-the-art lightweight architectures achieve real-time speed, this approach is not applicable to modern segmentation models for real-time speed up. Further, the approach sacrifices the segmentation accuracy of non-key frames even with accurate optical flow-based feature propagation, thus maintaining the accuracy of non-key frames in this pipeline still remains a problem to be addressed.~\citet{hu2020temporally} exploit temporal information in videos to tackle this problem via a temporally distributed network architecture. However, this strategy is suboptimal in that the sub-networks are designed independently of individual frames with no consideration of temporal locality across frames.

To tackle these issues, we propose a novel method to achieve practical speed-ups for real-time video semantic segmentation networks, exploiting spatial-temporal locality to create input-dependent sub-networks that preserves accuracy on both key and non-key frames at inference time. Specifically, we propose an input-dependent block-wise pruning mechanism to obtain inference time dynamic sub-networks tailored to each frame by exploiting the resiliency of residual networks to the removal of residual blocks at inference time due to their skip-connections. That is, we learn a masking mechanism that decides which residual blocks to drop and which to keep, conditioned on a given input frame and spatial-temporal information across previous frames. 

While the partial computation of residual blocks produces real-time speed-ups, it still suffers from the speed-accuracy trade-off problem. To compensate for this, we further exploit the temporal locality of video frames as depicted in Figure~\ref{fig:1}. In particular, we transform the features from previous frames for reuse in the static regions across adjacent frames. We do this by generating spatial masks that capture the distortion between adjacent frames. However in most video semantic segmentation datasets such as Cityscapes~\citep{cordts2016cityscapes}, not all frames are labelled hence we utilize knowledge distillation by employing a strong image segmentation model as a teacher network to train the spatial mask generator. 

\begin{wrapfigure}[16]{r}{0.45\textwidth}
    \small
    \centering
    \vspace{-0.22in}
    \includegraphics[width=0.46\textwidth]{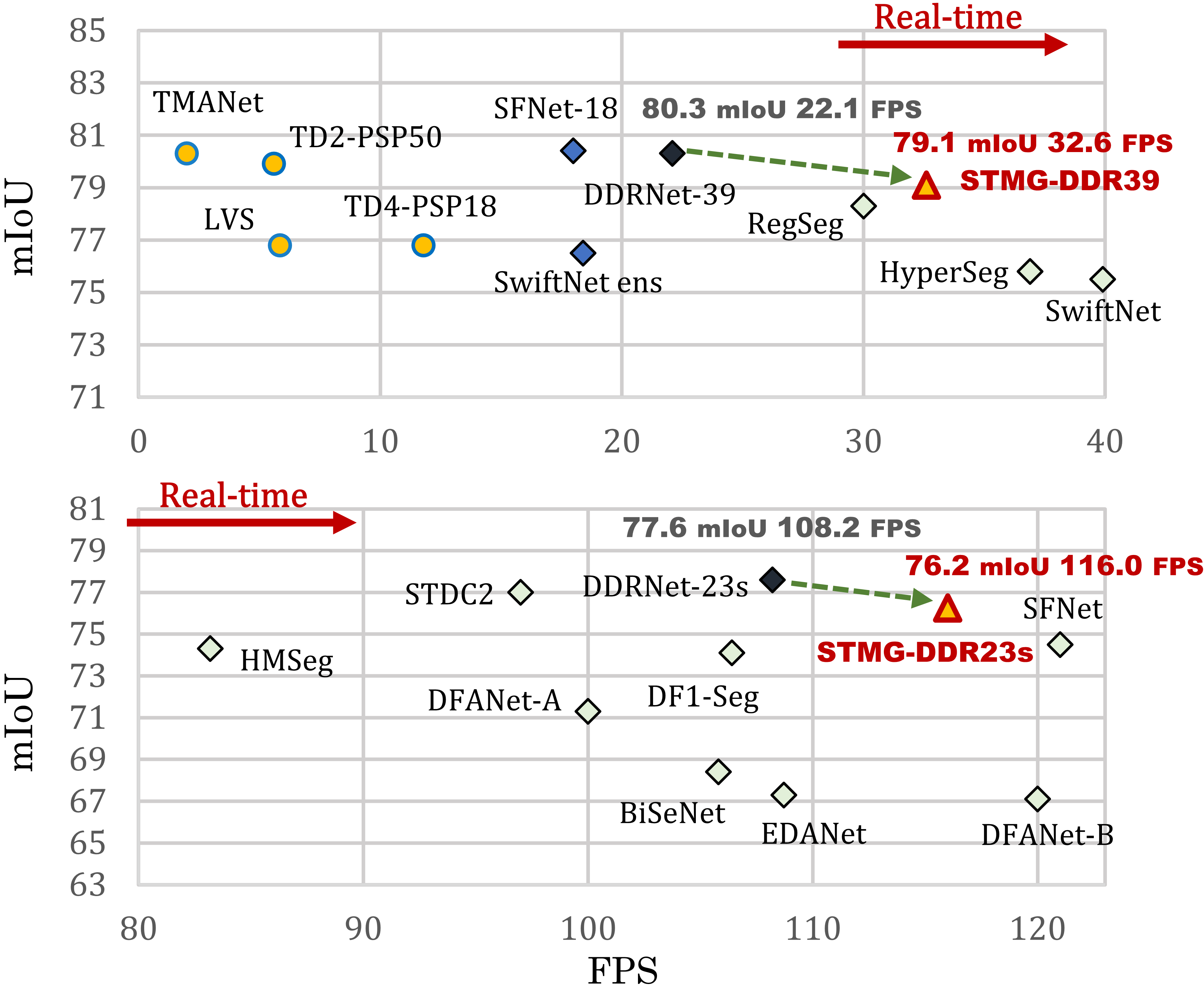}
    \vspace{-0.22in}
    \caption{\small \textbf{Performance on Cityscapes.} Our proposed method, Spatial-Temporal Mask Generator, provides a better trade-off between mIoU and FPS.}
    \label{fig:2}
\end{wrapfigure} 

In addition, to avoid information loss due to repeated partial computations for sequential frames, we introduce a distortion-aware scheduling policy which determines non-key frames to apply the proposed model. Specifically, we consider frames with distortions greater than some set threshold as key frames, and invoke the full network to compute their features. Crucially, when a frame is determined to be a non-key frame, we transform the feature from the previous frame, and perform partial computation of the backbone network only in spatial regions that change from the previous frame. The resulting model, Spatial-Temporal Mask Generator (STMG) achieves a better speed-accuracy trade-offs compared to semantic and video semantic segmentation models as shown in Figure~\ref{fig:2}. 
Our contributions are three fold:
\begin{itemize}[itemsep=0.5mm, parsep=1pt, leftmargin=*]
    \item {We propose a framework to speed up the inference of residual backbone networks for real-time video semantic segmentation by performing partial network evaluation and reusing temporally consistent features across frames.}
    \item{We propose a method to capture spatial-temporal information across frames based on inter-frame distortion without heavy optical flow computations.}
    \item{We validate our method on two benchmarks against multiple state-of-the-art real-time video segmentation models and show that our method can speed up models with residual backbones at inference time with marginal loss of accuracy.}
\end{itemize}

\begin{figure*}
    \vspace{-0.1in}
    \centering
        \includegraphics[width=\linewidth]{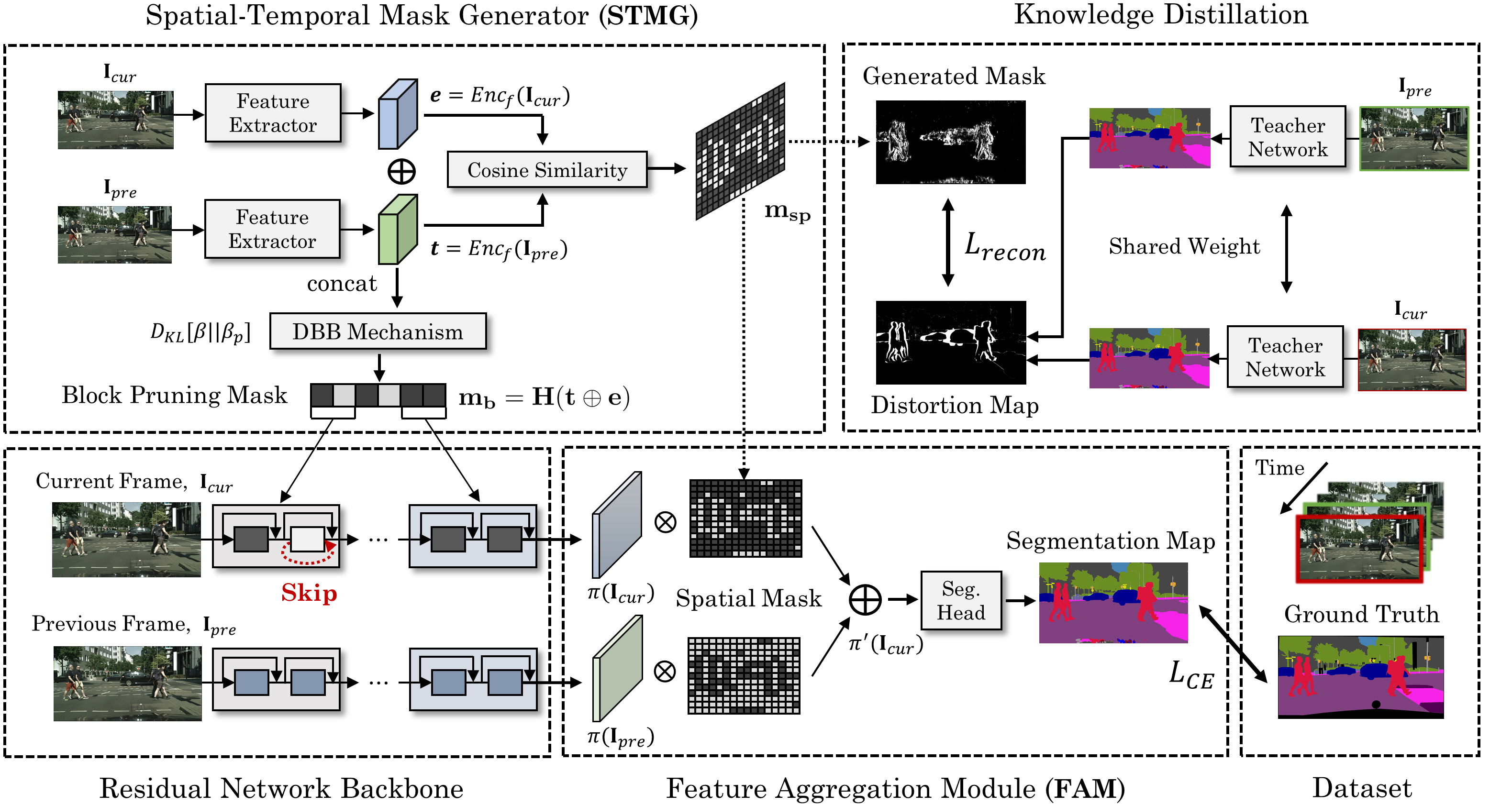}
    \vspace{-0.1in}
    \caption{\textbf{The overall process for video segmentation.} Our framework exploits spatial-temporal properties by extracting features from images of adjacent frames and utilizing them to compute both spatial mask and block pruning mask. In detail, a 2D mask is generated by calculating the cosine similarity between two features and utilized as a ratio to blend the features pixel-wisely (spatial mask). These spatial-temporal properties are transferred to learn the block dropping behavior by providing the extracted features to generate the block pruning mask by concatenation (block pruning mask).}
    \label{fig:3}
\end{figure*}
\section{Related Work} \label{related_work}

\paragraph{Semantic Segmentation}
While the literature on semantic segmentation is vast, we only describe few recent works that are relevant to ours. Deep convolutional neural networks have been widely used in semantic segmentation tasks by adopting a data-driven approach to improve accuracy. By providing an effective global contextual prior and fusing local and global context features, the networks can learn details and semantics simultaneously~\citep{zhao2017pyramid}. With urban scene properties where each row of the image has different statistics in terms of category distribution, each row of the feature map can be individually more focused on a specific channel using a height-driven attention map~\citep{choi2020cars}. Explicitly exploiting visual dependency relations, such as intra-class, inter-class and global dependence among semantic entities, networks can improve the generalization ability~\citep{liu2021exploit}.

\paragraph{Real-time Semantic Segmentation} Modern approaches to real-time semantic segmentation usually adopt a strategy of reducing the size of an input image provided to the residual backbone, as residual networks are considered to be bottlenecks in the overall computation~\citep{zhao2018icnet}. Reducing the size of an input image generally degrades spatial resolution, which leads to a degradation of performance. Hence recent approaches adopt a strategy of designing a dual-resolution path consisting of a spatial path and a context path~\citep{yu2018bisenet, yu2021bisenet}. The spatial path with a high-resolution feature adopts a small stride to preserve the spatial information while a context path with a low-resolution feature adopts a fast downsampling strategy to obtain sufficient receptive field. Bilateral connections between dual resolutions are continuously repeated as the network deepens for efficient information fusion to deliver the state-of-the-art performance of the latest models~\citep{hong2021deep}.

\paragraph{Video Semantic Segmentation}~\citet{zhu2017deep} and~\citet{li2018low} introduce scheduling policies to identify key frames to apply heavy neural networks and propagate the feature across multiple frames sequentially. That is, features are reused and this reduces the computational requirements on non-key frames. Adaptive scheduling policies that select key frames based on changing video dynamics instead of heuristic policies have been recently proposed in~\citet{xu2018dynamic},~\citet{wang2020dynamic} and~\citet{he2020temporal}. However, these policies are still suboptimal in that the candidates for sub-networks are limited to a subset of image segmentation network with feature propagation module which are pre-defined before the training stages resulting in a non-dynamic neural network architecture.

\paragraph{Dynamic Network Pruning} Residual networks~\citep{he2016deep} are composed of residual blocks and skip connections which enable them to behave like ensembles of relatively shallow networks~\citep{veit2016residual}. These networks have been shown to be resilient to layer dropping, usually with minimal loss of accuracy. In ~\citet{wang2018skipnet} and~\citet{wu2018blockdrop}, this resilience to layer dropping is exploited using reinforcement learning to prevent over-parameterization in residual networks.~\citet{li2020learning} exploits contextual information to develop dynamic routing methods that adapt to the scale of the input image for the semantic segmentation task. Furthermore,~\citet{he2021cap} focuses on utilizing spatial information for channel-wise pruning~\citep{he2017channel, lin2020channel} of the semantic segmentation networks. However, for video semantic segmentation, such methods that exploit spatial information are insufficient since they do not  consider the temporal nature of videos. Hence, to fully utilize contextual information in the video semantic segmentation task, the spatial-temporal properties have to be taken into account.

\paragraph{Dependent Beta-Bernoulli Process} In latent feature model, the Indian buffet process~\citep{ghahramani2005infinite}
defines a prior distribution on the binary feature indicators with potentially infinite number of features. Dependent Indian buffet process and its finite-dimensional beta-Bernoulli approximation~\citep{williamson2010dependent,zhou2011dependent,ren2011kernel} extend the Indian buffet process (IBP) to incorporate input covariates as follows:
\begin{align}
    \varphi_k \sim p(\varphi_k), \quad z_{n,k} | \varphi_k, \mathbf{x}_n \sim Ber( g(\varphi_k, \mathbf{x}_n) ),
\end{align}
where $g$ is an arbitrary function mapping $\varphi_k$ and $\mathbf{x}_n$ to a probability. The input covariates $\mathbf{x}_n$ corresponds to the inputs of a certain layer and dropout probabilities are adjusted according to $\mathbf{x}_n$ in the dependent beta-Bernoulli dropout. 

\section{Method} \label{method}
In this section, we first introduce our dynamic block pruning mechanism based on spatial-temporal locality. We then describe a spatial-temporal mask generator for feature transformation and reuse based on the distortion between adjacent frames. The overall framework consists of two networks, the segmentation network and the mask generator. Note that the proposed method is model-agnostic and applicable to any semantic segmentation model with a residual network.

\subsection{Input-dependent Block Pruning}

In this section, we describe the input-dependent block pruning mechanism. Let $\widehat{\mathbf{W}}$ be the parameters of a residual neural network with $\mathbf{x}_n$ as input and let $\mathbf{z}_n \in \{0, 1\}^K$ be a binary mask vector generated for the input $\mathbf{x}_n$ where $K$ is the number of prunable residual blocks. Note that the first block in each residual layer is not prunable. Conditioned on the input $\mathbf{x}_n$, we define the input-dependent residual block pruning probability $\boldsymbol{\varphi}(\mathbf{x}_n) \in [0,1]^K$ for $\mathbf{x}_n$ and generate a binary mask vector $\mathbf{z}_n$ for each block as follows:
\begin{equation}
    z_{n,k} \sim Ber(\varphi_k(\mathbf{x}_n)) \text{ for } k=1,\dots, K. 
\end{equation}

That is, given a residual network with $K$ prunable blocks, we model the block pruning probability of each individual block as a Bernoulli random process follows:
\begin{equation}
    \psi(\mathbf{u}|\mathbf{x}_n) = \prod_{k=1}^{K}\varphi_k(\mathbf{x}_n)^{u_{k}}(1-\varphi_k(\mathbf{x}_n))^{1-u_{k}},
\end{equation}
where $u_{k} \in \{0, 1\}$ indicates whether to prune or keep the $k$-th residual block.

In this paper, we model the transformation process from images into pruning probabilities with a Bayesian approach by inducing sparsity with the batch normalization~\citep{ioffe2015batch} layer of the pruning model inspired by~\citet{lee2018adaptive}. To obtain a sparse pruning mask, we impose a sparsity inducing prior on $\varphi_k(\mathbf{x}_n)$ and independently compute the residual block pruning probabilities as follows:
\begin{equation}
    \varphi_k(\mathbf{x}_n,\beta_k) = \mathrm{clamp}\bigg( \gamma_k \frac{g_k(\mathbf{x}_n)-\mu_k}{\sigma_k} + \beta_k, \tau\bigg),
\end{equation}
where $g_k(\mathbf{x}_n)$ denotes a transformation of $\mathbf{x}_n$ using the non-linear function $g$ for the $k$-th residual block, $\mu_k$ and $\sigma_k$ are the estimates of the $k$-th component's mean and standard deviation of the transformed inputs, $\gamma_k$ and $\beta_k$ are scaling and shifting parameters, and $\tau > 0$ is some tolerance to prevent overflow, which is set as $1e^{-10}$ while $\mathrm{clamp}(x, \tau) = \mathrm{min}(1-\tau, \mathrm{max}(\tau, x))$.

\begin{wrapfigure}[18]{l}{0.45\textwidth}
    \vspace{-0.05in}
    \small
    \centering
    \includegraphics[width=\linewidth]{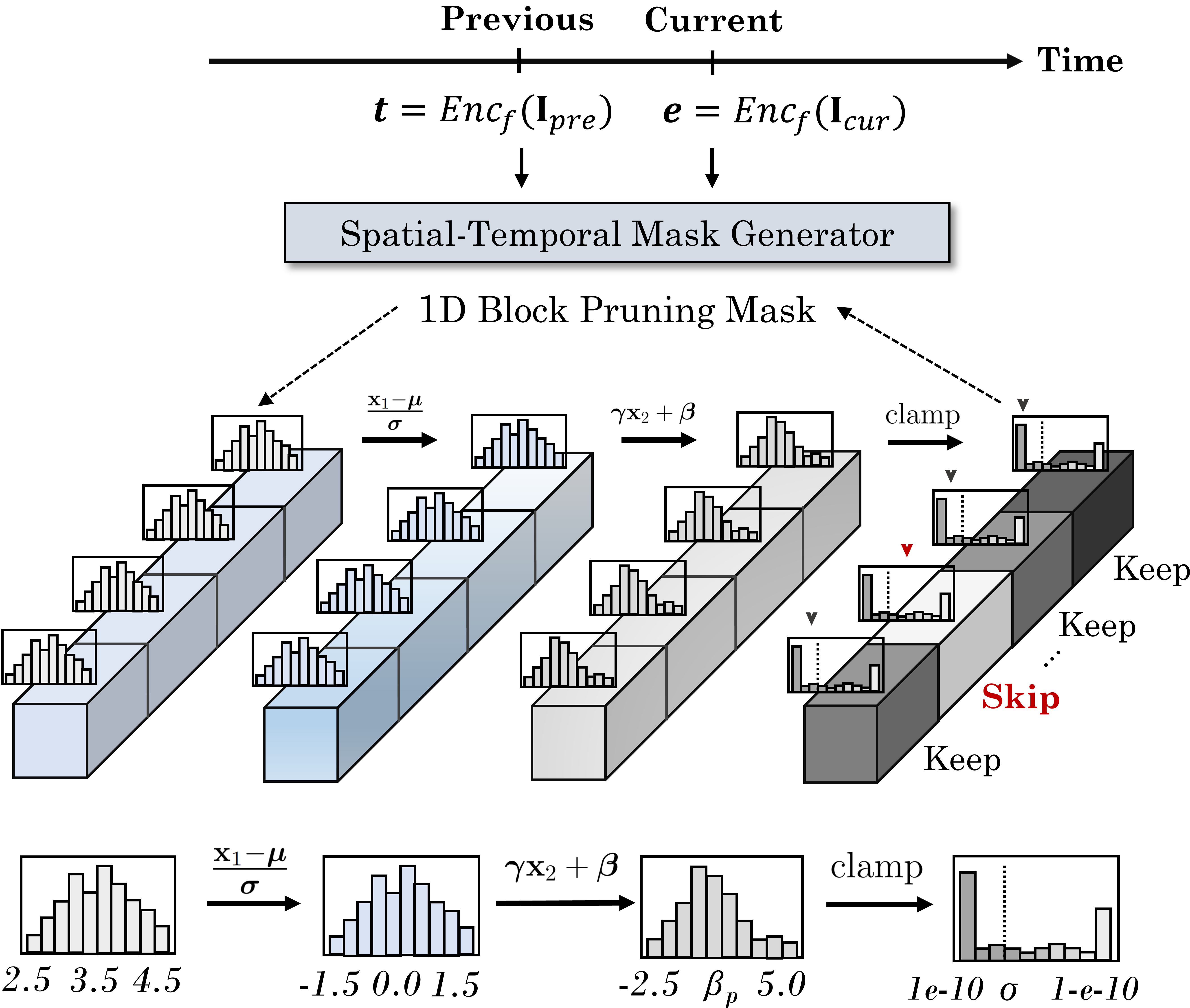}
    \vspace{-0.18in}
    \caption{\small \textbf{Illustration of input-dependent block pruning.} We visualize how the extracted features $\bf t$ and $\bf e$ are utilized by STMG for generating the residual block pruning probabilities.}
    \label{fig:4}
\end{wrapfigure}

During the optimization phase, $z_{n,k}$ is sampled from the relaxed Bernoulli distribution. Specifically, by passing $\boldsymbol{\delta}_{uc} \in \mathbb{R}^{K}$ to a softplus function and sampling $\boldsymbol{\epsilon} \sim \mathcal{N}(\mathbf{0},\mathbf{I})$ to provide a variability during the optimization stage, the sparsity inducing prior on $\boldsymbol{\varphi}(\mathbf{x}_n)$, which is $\mathcal{N}(\boldsymbol{\beta}_{p}, \rho \mathbf{I})$, and the trainable parameters of $\boldsymbol{\beta}$ are defined as follows:
\begin{equation}
    \boldsymbol{\beta} \sim \mathcal{N}(\boldsymbol{\beta}_{bn}, \boldsymbol{\delta}), \,\,
    \boldsymbol{\beta}_{p} \sim \mathcal{N}(\boldsymbol{\beta}_{p}, \rho \mathbf{I}),
\end{equation}
while $\boldsymbol{\beta}_{bn}$ is obtained with batch normalization layer and $\delta_{k} = \epsilon_{k} \cdot \mathrm{softplus}(\delta_{uc,k})$ is obtained with $\boldsymbol{\delta}_{uc}$ and $\boldsymbol{\epsilon}$ during the optimization phase.

That is, $D_{KL}[\boldsymbol{\beta} \Vert \boldsymbol{\beta}_{p}]$, the KL divergence loss for sparsity inducing prior on $\boldsymbol{\varphi}(\mathbf{x}_n)$ is defined as follows:
\begin{equation}
    D_{KL}[\boldsymbol{\beta} \Vert \boldsymbol{\beta}_{p}] = \log \frac{\rho}{\delta} + \frac{\delta^2 + (\boldsymbol{\beta}_{p}-\boldsymbol{\beta}_{bn})^2}{2\rho^2},
\end{equation}
where $\rho$ and $\boldsymbol{\beta}_{p}$ are constants for sparsity inducing prior and $\delta$ and $\boldsymbol{\beta}_{bn}$ are trainable parameters of the pruning mechanism. An illustration of the pruning method is depicted in Figure~\ref{fig:4}.

\subsection{Spatial-Temporal Mask Generator}
We describe the Spatial-Temporal Mask Generator (STMG) which generates two types of mask: a block pruning mask and a spatial mask, based on spatial-temporal information. An overview of the mask generator is depicted in Figure~\ref{fig:3}. 

\paragraph{Block Pruning Mask} Based on the input-dependent block pruning method previously described, we develop a block mask generator for prunable residual blocks. Specifically, the mask generator takes as input two adjacent frames, ${{\bf I}_{pre}}$ and ${{\bf I}_{cur}}$, and processes each frame using a feature extractor, $Enc_f$, with shared weights across all frames. We then concatenate and pass the outputs of the feature extractor to another convolution layer to generate a pruning mask with the same dimensionality as the number of residual blocks in the backbone network as follows:
\begin{equation}
    {\bf t} = Enc_f ({{\bf I}_{pre}}), \;\, {\bf e} = Enc_f ({{\bf I}_{cur}}), \:\, {\bf m_{b}} = {\bf {H}} ({\bf t} \oplus {\bf e}),
\end{equation}
where $\bf t$ and $\bf e$ are extracted features from the previous and current frame, $\bf m_{b}$ denotes a block pruning mask and $\bf H$ denotes overall transformation of the input in the pruning mechanism.

\paragraph{Spatial Mask} In addition, to compensate for information loss due to the partial computation of residual blocks, we propose a spatial masking scheme which exploits the temporal locality of videos. Specifically, we generate a spatial mask, ${\bf m_{sp}}$, which predicts unchanged regions between adjacent frames together with the block pruning mask. By applying the spatial mask on the features of the previous frame, we transform the feature from the previous frame for reuse in the current frame. This is done by computing the cosine similarity between the outputs of the feature extractor from the previous frame and current frame denoted as $\bf t$ and $\bf e$ as follows:
\begin{equation}
    \textit{m}_{sp,i} ({\bf t},{\bf e}) = -0.5 \times cos(t_i, e_i) + 0.5,
\end{equation}
where $\textit{m}_{sp,i}$ denotes a spatial mask of the $i$-th pixel and we scale the value to be in the range of $[0,1]$.

\paragraph{Knowledge Distillation} However, the cosine similarity between ${\bf t}$ and ${\bf e}$ from adjacent frames alone is not sufficient for encoding meaningful spatial-temporal information. To further improve the quality of the features, we add an auxiliary loss to match the generated spatial mask with the ground-truth distortion map. Since not all frames in the dataset are labeled, we utilize knowledge distillation~\citep{hinton2015distilling} to generate ground-truth distortion map as shown in Figure~\ref{fig:6} (b). Specifically, we generate the ground-truth distortion map by subtracting the segmentation maps obtained by applying a strong image segmentation model on two adjacent frames. Following \citet{deng2018learning}, the loss function for spatial mask consists of the binary cross-entropy loss and dice loss, as follows:
\begin{equation}
    L_{recon}({\bf m_{sp}}, {\bf n}) = L_{bce}({\bf m_{sp}}, {\bf n}) + L_{dice}({\bf m_{sp}}, {\bf n}),
\end{equation}
where $\bf n$ denotes the ground-truth distortion map obtained by knowledge distillation. The dice loss alleviates the class-imbalance problem, as only partial spatial areas are distorted between adjacent frames. The dice loss, $L_{dice}$, is computed as follows:
\begin{equation}
    L_{dice} = 1 - \frac{2 \sum{m_{sp,i}n_{i}} + \kappa}{\sum{m_{sp,i}^2} + \sum{n_{i}^2} + \kappa},
\end{equation}
where $\kappa$ is a smoothing value set as 1 to avoid zero division. $L_{dice}$ is the cornerstone for generating crisp edges while optimizing only the cross-entropy loss leads to relatively coarse results.

\paragraph{Feature Aggregation} Using the generated spatial mask, we aggregate features of the segmentation model from the previous frame and current frame as follows:
\begin{equation}
    \pi^{\prime}({{\bf I}_{cur}}) = {\bf m_{sp}} \cdot \pi({{\bf I}_{cur}}) + (\mathbf{I} - {\bf m_{sp}}) \cdot \pi({{\bf I}_{pre}}),
\end{equation}
where $\pi(\bf{\cdot})$ is the output of the backbone segmentation network.
In particular, we intend the network to focus more on features for distorted regions, while reusing features from previous frames for static regions. Specifically, we pass the aggregated feature $\pi^{\prime}({{\bf I}_{cur}})$ to the segmentation head module to obtain the semantic segmentation map for the current frame. 

\begin{figure}[t]
    \begin{minipage}{0.6\textwidth}
        \includegraphics[width=\linewidth]{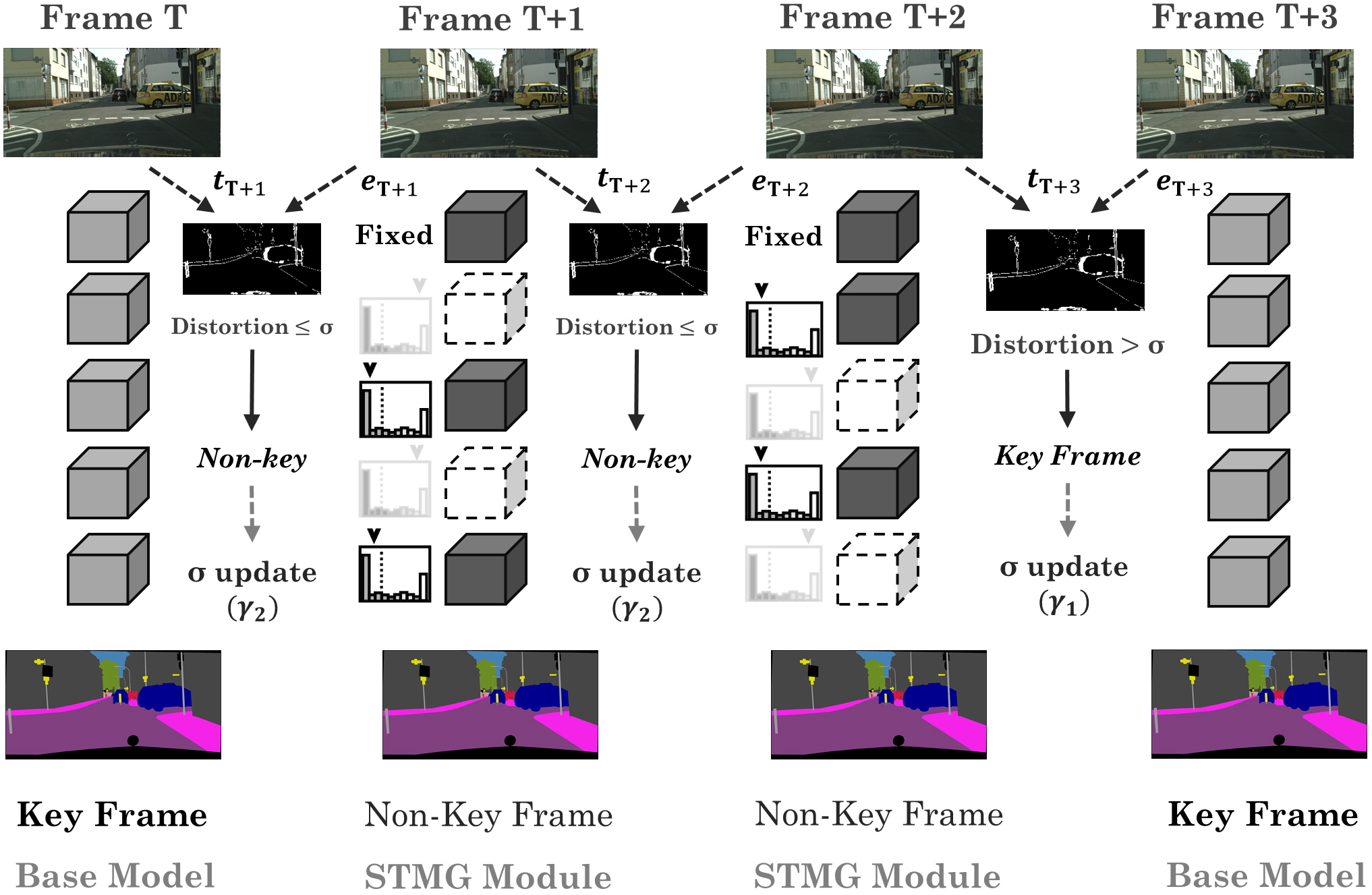}
    \end{minipage}
    \hspace{0.05in}
    \begin{minipage}{0.39\textwidth}
        \centering
        \footnotesize
        \input{table/PseudoCode}
    \end{minipage}
    \caption{\textbf{(Left)}: Illustration of distortion-aware key frame selection. At the arrival of a new frame, we first calculate the distortion of adjacent frames to determine key and non-key frames. \textbf{(Right)}: Pseudo-code for distortion-aware scheduling policy. For good balance between key and non-key frames, we update the threshold value scaled by a factor $\gamma$.}
    \label{fig:5}
    \vspace{-0.2in}
\end{figure}

\subsection{Distortion-Aware Scheduling Policy}
While we compensate for information loss by reusing the features from previous frames, error in the features propagate as time progresses. Hence, we propose a scheduling policy that applies the proposed module only to specific non-key frames. We do this by capturing the amount of distortion using the spatial mask for each frame. Specifically, if the distortion for a given frame is greater than some set threshold, we consider it as a key frame and compute the full network on the frame. However, if the distortion is lower than the threshold, we consider the frame a non-key frame and transform the features from the previous frame for reuse as well as perform partial computation of the residual blocks using the generated input-dependent mask.

\paragraph{Scaling Factor}
After each computation, we update the threshold value with the distortion amount of the current frame scaled by a factor $\gamma$. To prevent the full network from being applied to more than one frame in a row, we assign a value of 2 to $\gamma_1$ and update the threshold value with a multiplicative factor of $\gamma_1$ right after a key frame so that the next frame can be determined as a non-key frame. Additionally, to avoid situations where partial computations are repeated over sequential frames, we assign a value of 0.95 to $\gamma_2$ right after a non-key frame and update the threshold value with the current distortion factor multiplied by $\gamma_2$ so that the next frame can be slightly induced as a key frame. Most video semantic segmentation frameworks adopt a strategy of propagating features across multiple frames sequentially, however this require a heavy optical flow computation since errors propagate as time progresses. Since our method targets real-time speed-up without optical flow computation, we adopt a strategy of frequently switching between key and non-key frames with proposed scaling rules.

These scaling rules provide a good balance between key and non-key frames while simultaneously capturing video dynamics and leveraging spatial-temporal locality. An illustration of this scheduling policy together with pseudo-code is provided in Figure~\ref{fig:5}.

\begin{figure}[!t]
    \begin{minipage}{0.5\linewidth}
        \centering
        \captionof{table}{\small Performance comparison with \textbf{real-time (> 30 FPS)} semantic segmentation models on Cityscapes~\citep{cordts2016cityscapes}.}
        \vspace{-0.06in}        
        \resizebox{1\textwidth}{!}{
        \renewcommand{\arraystretch}{0.98}
        \renewcommand{\tabcolsep}{1.0mm}
        \begin{tabular}{lcccccccccccc}
        \toprule
        Model & mIoU & FPS & FLOPs & Params \\
        \midrule
        PP-LiteSeg-T1~\citep{peng2022pp} & 73.1 & \textbf{273.6} & - & - \\
        SwiftNetRN-18~\citep{orsic2019defense} & 75.4 & 39.9 & 104.0G & 11.8M \\
        SFNet-DF2~\citep{li2020semantic} & 75.8 & 61.0 & 48.5G & 10.5M \\
        BiSeNetV2-L~\citep{yu2021bisenet} & 75.8 & 47.3 & 118.5G & - \\
        HyperSeg~\citep{nirkin2021hyperseg} & 76.2 & 36.9 & 7.5G & 10.1M \\
        CABiNet~\citep{kumaar2021cabinet} & 76.6 & 76.5 & 12.0G & 2.6M \\
        STDC2-Seg75~\citep{fan2021rethinking} & 77.0 & 97.0 & - & - \\
        RegSeg~\citep{gao2021rethink} & 78.1 & 30.0 & 39.1G & 3.3M \\
        \midrule
        DDRNet-39~\citep{hong2021deep} & 80.3 & 22.1 & 261.8G & 32.4M \\
        \textbf{- STMG-DDR39} & \textbf{79.1} & 32.6 & 175.2G & 22.9M \\
        \midrule
        DDRNet-23-Slim~\citep{hong2021deep} & 77.6 & 108.2 & 33.8G & 5.7M \\
        \textbf{- STMG-DDR23-Slim} & 76.2 & 116.0 & 27.2G & 4.9M \\
        \bottomrule
        \end{tabular}}
        \label{tab:cityscapes_small}
    \end{minipage}
    \hfill
    \begin{minipage}{0.47\linewidth}
        \centering
        \captionof{table}{\small Performance comparison with video semantic segmentation models on Cityscapes~\citep{cordts2016cityscapes}.}
        \vspace{-0.06in}
        \resizebox{1\textwidth}{!}{
        \renewcommand{\arraystretch}{0.93}
        \renewcommand{\tabcolsep}{1.0mm}
        \begin{tabular}{lcccc}
        \toprule
        Model & mIoU & FPS & Max Latency  \\
        \midrule
        DVSNet~\citep{xu2018dynamic} & 63.2 & 30.3 & - \\
        CLK~\citep{shelhamer2016clockwork} & 64.4 & 6.3 & 198 (ms) \\
        DFF~\citep{zhu2017deep} & 69.2 & 6.4 & 575 (ms) \\
        GRFP(5)~\citep{nilsson2018semantic} & 73.6 & 3.9 & 255 (ms) \\
        TD4-Bise18~\citep{hu2020temporally} & 75.0 & 47.6 & 21 (ms) \\
        FANet18~\citep{hu2020real} & 75.5 & 72.0 & 14 (ms) \\
        PEARL~\citep{jin2017video} & 76.5 & 1.3 & 800 (ms) \\
        LVS~\citep{li2018low} & 76.8 & 5.8 & 380 (ms) \\
        TD2-PSP50~\citep{hu2020temporally} & 79.9 & 5.6 & 178 (ms) \\
        TMANet50~\citep{wang2021temporal} & 80.3 & 2.0 & 500 (ms) \\
        NetWarp~\citep{gadde2017semantic} & \textbf{80.6} & 0.3 & 3004 (ms) \\
        \midrule
        \textbf{STMG-DDR39} & 79.1 & 32.6 & 46 (ms) \\
        \textbf{STMG-DDR23-Slim} & 76.2 & \textbf{116.0} & \textbf{10 (ms)} \\    
        \bottomrule
        \end{tabular}}
        \label{tab:cityscapes_video}
    \end{minipage}
    \vspace{-0.2in}
\end{figure}
\section{Experiments} \label{experiment}
\subsection{Setup and Implementation}
\paragraph{Dataset} We evaluate our method on Cityscapes~\citep{cordts2016cityscapes}. Cityscapes is a dataset for urban-scene parsing consisting of 5,000 images of urban street scenes with 19 classes. The dataset contains 2,975 finely annotated images for training, 500 images for validation, and 1,525 images for testing. We evaluate our method based on the standard semantic segmentation evaluation metrics: mean class-wise intersection over union (mIoU), and Frames Per Second (FPS). Cityscapes is made up of snippets each consisting of 30 frames with the 20th frame of each snippet annotated. Hence, we train and evaluate our model on the 20th frame of each snippet with the image from the previous frame given. We also evaluate our method on CamVid~\citep{brostow2009semantic} which is a road-scene parsing dataset. The results for the CamVid experiments are reported in Section~\ref{sup/camvid} of the supplementary file.

\paragraph{Models and Baselines} We select two state-of-the-art real-time semantic segmentation models for our experiments: DDRNet-39 and DDRNet-23-Slim. Since DDRNet models have residual blocks both in the high and low resolution branches, we apply our dynamic block pruning mechanism in both branches. To preserve the output dimension of each residual layer, we do not drop the first residual block in each layer. This is because dropping the first block in each layer will not allow forward propagation of the features in residual networks. Specifically, there are a total of 6 and 17 prunable residual blocks in DDRNet-23-Slim and DDRNet-39 respectively. 

\paragraph{Implementation} To train our models, we use mini-batch stochastic gradient descent (SGD) with weight decay of 0.0005 and the momentum of 0.9. We use a batch size of 16 for training on the Cityscapes dataset. We set the initial learning rate as 0.01 and train the model for 1,000 epochs for DDRNet-23-Slim and 484 epochs for DDRNet-39. We also apply random cropping for data augmentation with crop size of 1024 × 1024. We implement all models and conduct all experiments using PyTorch. We train each model with two GeForce RTX 3090 GPUs for DDRNet-39 base models, and two GeForce RTX 2080 Ti GPUs for DDRNet-23-Slim base models to reproduce the results.

\paragraph{Inference Speed Measurement} 
We conduct inference speed measurement on a single RTX 2080 Ti. Following the evaluation procedure of~\citet{si2019real},~\citet{orsic2019defense}, and~\citet{hong2021deep}, we remove the batch normalization layers after each convolutional layer during the speed measurement. Specifically, we use the protocols established by~\citet{chen2019fasterseg} for a fair comparison. Since not all frames are annotated in Cityscapes, we evaluate the speed of our model by sampling a portion of the entire video sequence to include both key and non-key frames in the annotated frames. Further, we report the maximum latency of our model along with the speed of our model as video semantic segmentation models have varying latency between key and non-key frames. We compare the speed of our model on two semantic segmentation backbone architectures: DDR-39 and DDR-23-Slim.

\begin{figure*}[t]
    \centering
        \includegraphics[width=\linewidth]{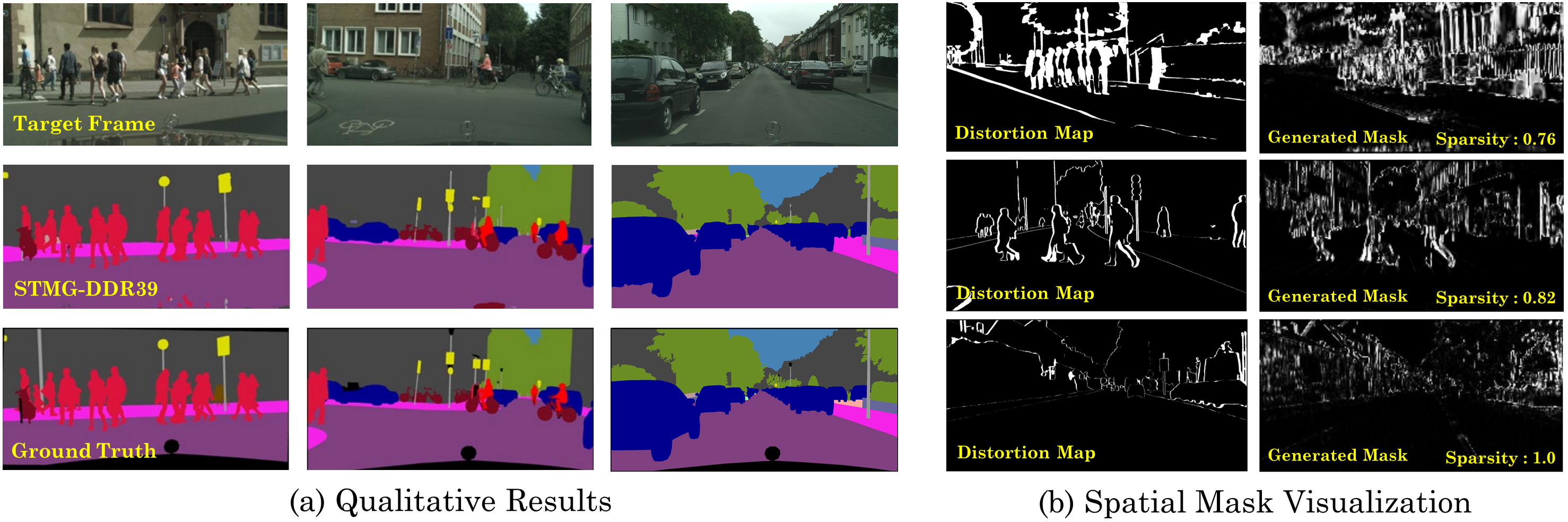}
    \vspace{-0.15in}
    \caption{\textbf{Qualitative results} of (a) our method and  (b) visualization of the spatial mask. Black and white represent static and distorted regions, respectively. During the training phase, a distortion map is provided to STMG to better capture the spatial-temporal locality between adjacent frames. The degree of sparsity refers to the pruning ratio among prunable residual blocks in the backbone.}
    \vspace{-0.12in}
    \label{fig:6}
\end{figure*}

\begin{figure}[!t]
    \vspace{-0.05in}
    \begin{minipage}{0.58\linewidth}
        \vspace{0.1in}
        \centering
        \captionof{table}{\footnotesize \textbf{Ablation study} on spatial and block pruning mask. $m_{sp}$ denotes a blending ratio of previous and current features while $m_{avg}$ is the average value of $\bf m_{sp}$ in the corresponding STMG.}
        \vspace{-0.16in}
        \resizebox{1\textwidth}{!}{
        \renewcommand{\arraystretch}{0.94}
        \renewcommand{\tabcolsep}{1.0mm}
    
        \begin{tabular}{lcccc} \\
        \toprule
        Pruning and Masking Methodology & mIoU & FLOPs & Params \\
        \midrule
        Random Block Pruning \footnotesize{(w/o} \textit{ft}) & 71.40 $\pm$ 0.14 & 175.6 $\pm$ 0.4G & 22.8M \\
        Random Spatial Masking & 78.06 $\pm$ 0.02 & 175.2G & 22.9M \\
        Uniform Blending ($m_{sp}=m_{avg}$) & 78.69 & 175.2G & 22.9M \\
        Uniform Blending ($m_{sp}=0.5$) & 78.25 & 175.2G & 22.9M \\
        \midrule
        \textbf{Ours (Spatial-Temporal Masking)} & \textbf{79.09} & \textbf{175.2G} & 22.9M \\
        \bottomrule
        \end{tabular}}
        \label{tab:abl1}
        \vspace{-0.02in}

        \captionof{table}{\footnotesize \textbf{Ablation study} on distortion-aware scheduling policy. $R$ denotes a key frame duration length for the fixed policy and $\gamma$ denotes a threshold scaling factor for the distortion-aware policy.}
        \vspace{-0.08in}
        \resizebox{1\textwidth}{!}{
        \renewcommand{\arraystretch}{0.9}
        \renewcommand{\tabcolsep}{1.0mm}
        \begin{tabular}{lcccc} \\
        \toprule
        Scheduling Policy & mIoU & FLOPs & Params \\
        \midrule
        Distortion-Aware Scheduling ($\gamma=0.95$) & 79.09 & \textbf{175.2G} & \textbf{22.9M} \\
        Distortion-Aware Scheduling ($\gamma=0.9$) & 79.43 & 187.6G & 24.2M \\
        \midrule
        Fixed Scheduling ($R=2$) & \textbf{79.73} $\pm$ 0.03 & 189.9G & 24.5M \\
        \bottomrule
        \end{tabular}}
        \label{tab:abl2}
        \vspace{+0.09in}
    \end{minipage}
    \hfill
    \begin{minipage}{0.39\linewidth}
        \centering
        \captionof{table}{\footnotesize FPS gain results. For a fair comparison, we select and compare the model and framework whose FPS of the base network is closest to the base network of ours.}
        \vspace{-0.08in}        
        \resizebox{1\textwidth}{!}{
        \renewcommand{\arraystretch}{1.08}
        \renewcommand{\tabcolsep}{1.0mm}
        
        \begin{tabular}{lcccc}
            \toprule 
            \centering
            Framework (Model) & FPS gain & mIoU drop \\
            \midrule
            TD4-Bise18 (Bise34)~\citep{hu2020temporally} & 28.57\% & 1.32\% \\ 
            \textbf{STMG-DDR39 (DDR39)} & \textbf{47.51}\% & 1.47\% \\
            \bottomrule
    	\end{tabular}}
        \label{tab:framework}
        \vspace{-0.02in}

        \captionof{table}{\footnotesize \textbf{Ablation study} on FAM. For a fair comparison, we use the same sparsity inducing prior $\beta_{p}$ and the same KL scale factor for per-frame block-wise pruning inspired by dependent beta-Bernoulli dropout~\citep{lee2018adaptive}.}
        \vspace{0.01in}
        \resizebox{1\textwidth}{!}{
        \renewcommand{\arraystretch}{1.04}
        \renewcommand{\tabcolsep}{1.0mm}
        \begin{tabular}{lcccc}
        \toprule
        Model & mIoU & Drop & FPS & Gain \\
        \midrule
        DDRNet-39 & 80.27 & - & 22.1 & - \\
        - Per-Frame Pruning & 76.31 & 4.93\% & 41.7 & 88.7\% \\
        - \textbf{Ours (STMG-DDR39)} & \textbf{79.09} & 1.47\% & 32.6 & 47.5\% \\
        \bottomrule
        \end{tabular}}
        \label{tab:abl3}
    \end{minipage}
    \vspace{-0.20in}
\end{figure}

\subsection{Results}
In Table~\ref{tab:cityscapes_small}, we report the experimental results on Cityscapes for the DDRNet-39 backbone network. The results show that our method reaches 32.6 FPS which is a 47.5\% gain over the base network which runs at 22.1 FPS on high resolution images. The segmentation outputs are shown in Figure~\ref{fig:6} (a) and in Section~\ref{sup/cityscapes} of the supplementary file. We report similar results for the DDRNet-23-Slim architecture in Table~\ref{tab:cityscapes_small}. As shown in Table~\ref{tab:cityscapes_small}, our method reaches 116.0 FPS with 7.2\% FPS gain compared to the base network. In both architectures, we increase the FPS with less than 2\% drop in accuracy. That is, STMG provides a better trade-off between accuracy and speed compared to state-of-the-art models. We further visualized this speed-accuracy trade-off in Figure~\ref{fig:2}.

\begin{wrapfigure}[16]{r}{0.305\textwidth}
    \vspace{-0.18in}
    \captionof{table}{\scriptsize Performance on Cityscapes over the DDRNet-39. "+dbb-DBB" represents a pruning mechanism using the proposed distortion bias-based DBB (dbb-DBB). STMG uses a default DBB~\citep{lee2018adaptive}-inspired pruning.}
    \vspace{-0.16in}
    \resizebox{0.3\textwidth}{!}{
    \renewcommand{\arraystretch}{0.98}
    \renewcommand{\tabcolsep}{1.0mm}
    \begin{tabular}{lcccc} \\
    \toprule
    Model & mIoU & FLOPs & Params \\
    \midrule
    STMG & 79.1 & \textbf{175.2G} & \textbf{22.9M} \\
    STMG + dbb-DBB & \textbf{79.3} & 188.7G & 24.8M \\
    \bottomrule
    \end{tabular}}
    \label{tab:dbbdbb}
    \vspace{0.1in}
    \small
    \centering
    \includegraphics[width=0.3\textwidth]{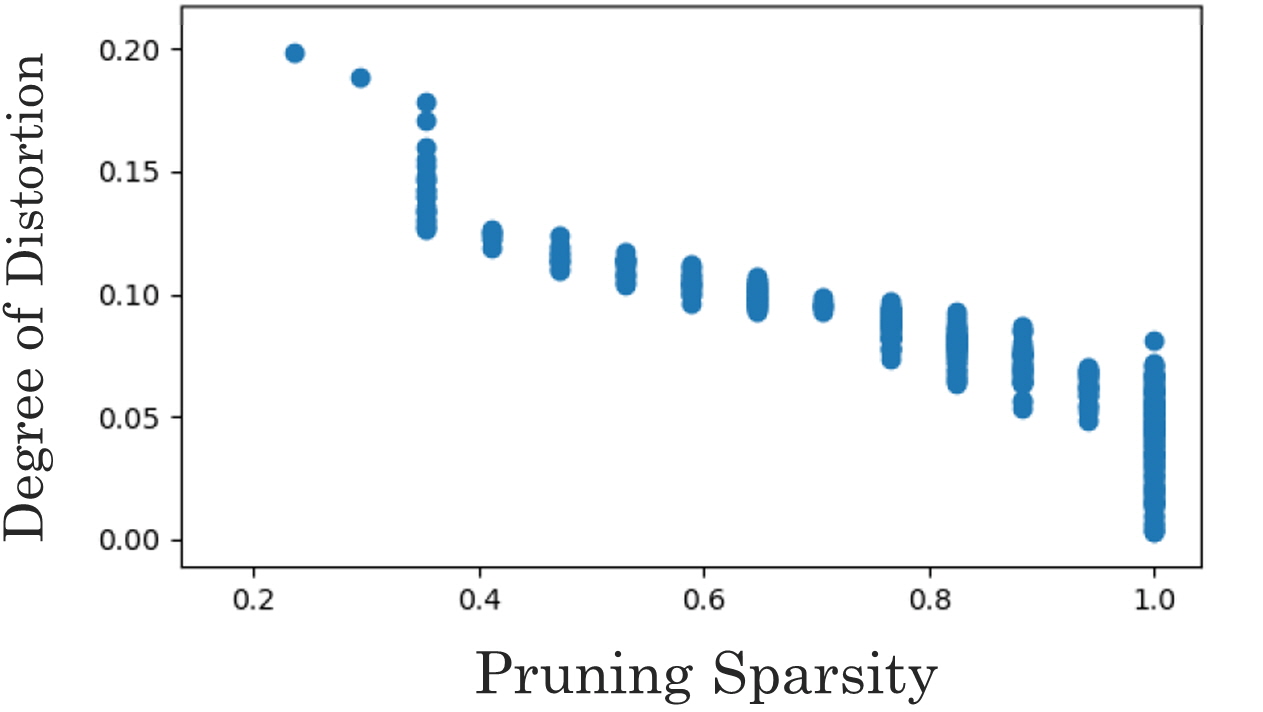}
    \vspace{-0.23in}
    \captionof{figure}{\scriptsize Correlation between the degree of distortion estimated by the spatial mask generator and pruning sparsity in dbb-DBB.}
    \label{fig:8}
\end{wrapfigure}

\paragraph{Degree of Distortion and Sparsity} We visualize the generated spatial masks in Figure~\ref{fig:6} (b). We observe that the mask generator successfully encodes meaningful spatial-temporal information and captures the distorted regions such as the edge movement of people or bicycles. As shown in Figure~\ref{fig:6} (b), each example shows different degrees of sparsity and the results show a strong correlation between sparsity and degree of distortion. To further improve the encoding of spatial-temporal information, we add a distortion bias $\eta$ to DBB~\citep{lee2018adaptive}-inspired block pruning mechanism as follows:
\begin{equation}
    \varphi_k(\mathbf{x}_n,\beta_k) = \mathrm{clamp}\bigg( \gamma_k  \frac{g_k(\mathbf{x}_n) + \eta -\mu_k}{\sigma_k} + \beta_k, \tau\bigg), \;\; \eta=-\overline{m_{sp,i}}+0.5,
\end{equation}
where $\varphi_k(\cdot)$ denotes a block pruning probability of the $k$-th residual block. The resulting pruning mechanism, distrotion bias-based DBB (dbb-DBB) achieves a stronger correlation between sparsity and degree of distortion as shown in Figure~\ref{fig:8}. The comparison results are reported in Table~\ref{tab:dbbdbb}.

\begin{figure*}[t]
    \centering
        \includegraphics[width=\linewidth]{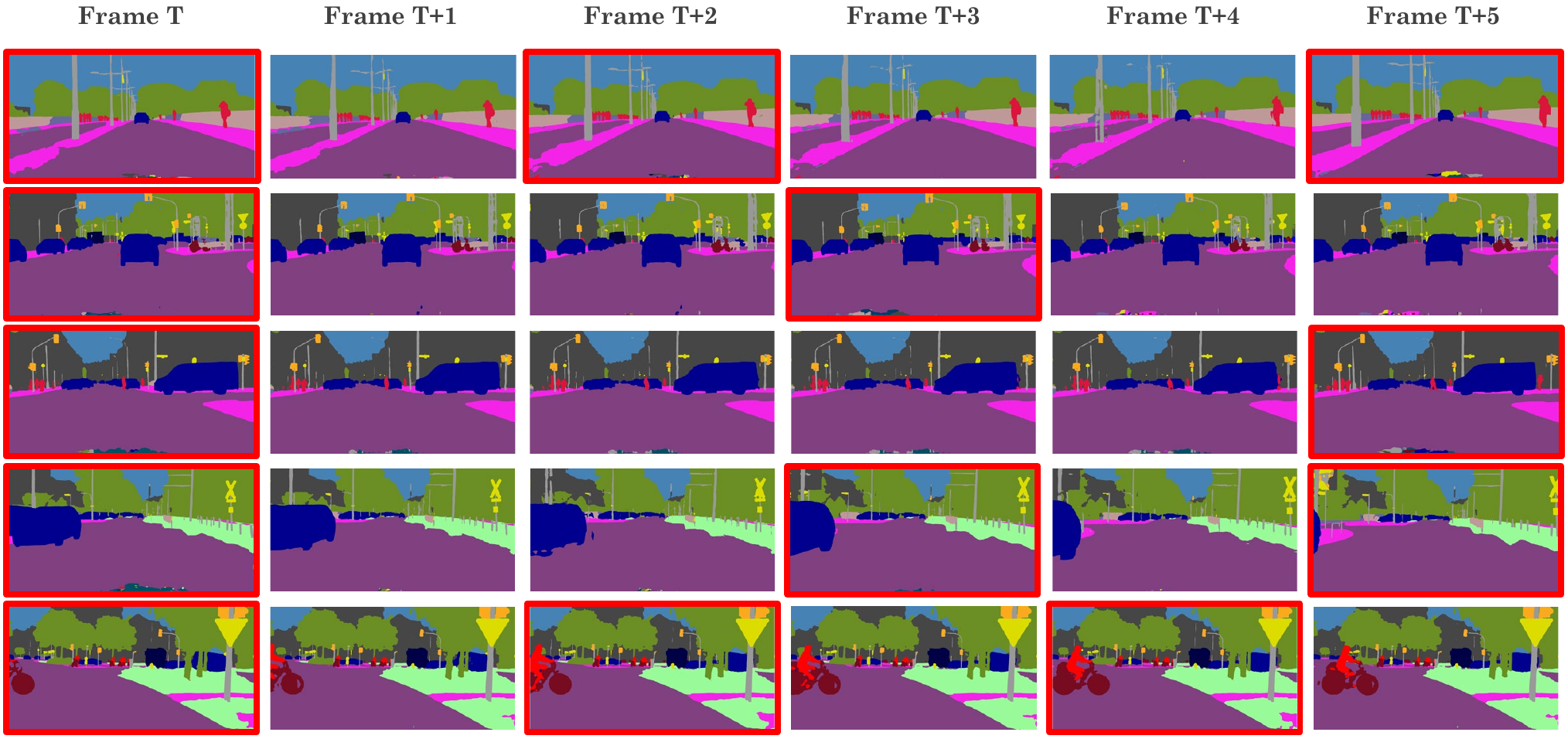}
    \vspace{-0.15in}
    \caption{\textbf{Qualitative results of distortion-aware scheduling policy.} Frames highlighted in red represent key frames and other frames represent non-key frames. The third row shows that for static video streams, the scheduling policy continues to perform partial computations on non-key frames.}
    \label{fig:7}
    \vspace{-0.2in}
\end{figure*}

\paragraph{Distortion-Aware Scheduling Policy} As shown in Figure~\ref{fig:7}, the adaptive scheduling policy successfully determines key and non-key frames by utilizing the distortion between adjacent frames. Diverse dynamic key frame paths are generated which shows that the scheduling policy does not generate a fixed configuration of key and non-key frames. For example, the 3rd row shows a repeated non-key frame as the distortion over consecutive frames are small, and the 5th row shows repeated switching between key and non-key frame when the distortion across frames becomes large.

\paragraph{Ablation Studies} We conduct ablation studies to examine the effectiveness of each component in our framework. As described in Table~\ref{tab:abl1}, we conduct an ablation study on spatial masking and block pruning. The results show that our proposed spatial masking trained with knowledge distillation outperforms the model that uses random or uniform blending with features of previous frame. The masking method differs during the inference time and the identical trained weights are used during the ablation study. Further, our DBB~\citep{lee2018adaptive}-inspired pruning mechanism outperforms the random pruning at the same computational cost. As described in Table~\ref{tab:abl2}, the results show that our distortion-aware scheduling policy achieves lower FLOPs and parameters compared to the fixed scheduling policy. In Table~\ref{tab:abl3}, we conduct an ablation study on our proposed feature aggregation module (FAM), and we observe that the information from the previous frame can compensate for the information loss due to the dropped blocks in the backbone for the current frame.

We discuss the limitations and societal impacts of our work in the supplementary file, in Section~\ref{sup/discussion}.
\section{Conclusion} \label{conclusion}
In this work, we proposed an efficient framework for real-time video semantic segmentation that exploits the spatial-temporal locality of videos. Our framework reuses the features from previous frames, and perform partial evaluation of the backbone network by taking into account the differences between two consecutive frames. This partial network evaluation is done using an input-dependent gating mechanism that decides which features and blocks to prune. Our framework is model-agnostic and can be applied to any semantic segmentation models with residual backbone networks to speed up their inference. We validated our method on two benchmark datasets with different semantic segmentation models and showed that it significantly improves inference-time speed at the expense of marginal drop in accuracy compared to baselines approaches that achieve lower inference speed-ups with similar decreases in model accuracy.

\clearpage

{\large
\bf
Acknowledgements
\vspace{+0.1in}
}
\\
This research was supported by Institute of Information \& communications Technology Planning \& Evaluation (IITP) grant funded by the Korea government(MSIT)  (No.2019-0-00075, Artificial Intelligence Graduate School Program(KAIST)), and Service Development and Demonstration on DNA Drone Technologies Program through the National Research Foundation of Korea(NRF) funded by the Ministry of Science and ICT(1711119697).

\bibliography{reference}

\clearpage
\appendix

\paragraph{
\begin{center}
\Large
Supplementary File
\vspace{+0.03in}
\end{center}
}

\paragraph{Organization}
The supplementary file is organized as follows. In Section~\ref{sup/camvid}, we first describe the experimental results on CamVid~\citep{brostow2009semantic} of the proposed STMG framework and describe the implementation detail and spatial masking method on CamVid. Also, we provide the code and dataset in Section~\ref{sup/code}. Then, we provide additional experimental results on Cityscapes~\citep{cordts2016cityscapes} with visualization of examples in Section~\ref{sup/cityscapes}. Finally, we provide the additional experimental results on CamVid in Section~\ref{sup/camvidadd} and potential negative societal impacts and limitation of our work in Section~\ref{sup/discussion}.

\section{Experiments on CamVid \label{sup/camvid}}
\vspace{-0.05in}
\paragraph{CamVid Dataset} 

CamVid~\citep{brostow2009semantic} is a road-scene parsing dataset which is taken from
the point of view of a driving automobile. The dataset consists of 701 annotated images with 11 classes at 1 Hz and in part 15Hz, in which 367, 101, and 233 images for training, validation, and testing, respectively.

\begin{figure}[h]
    \begin{minipage}{0.4\linewidth}
        \centering
            \includegraphics[width=\linewidth]{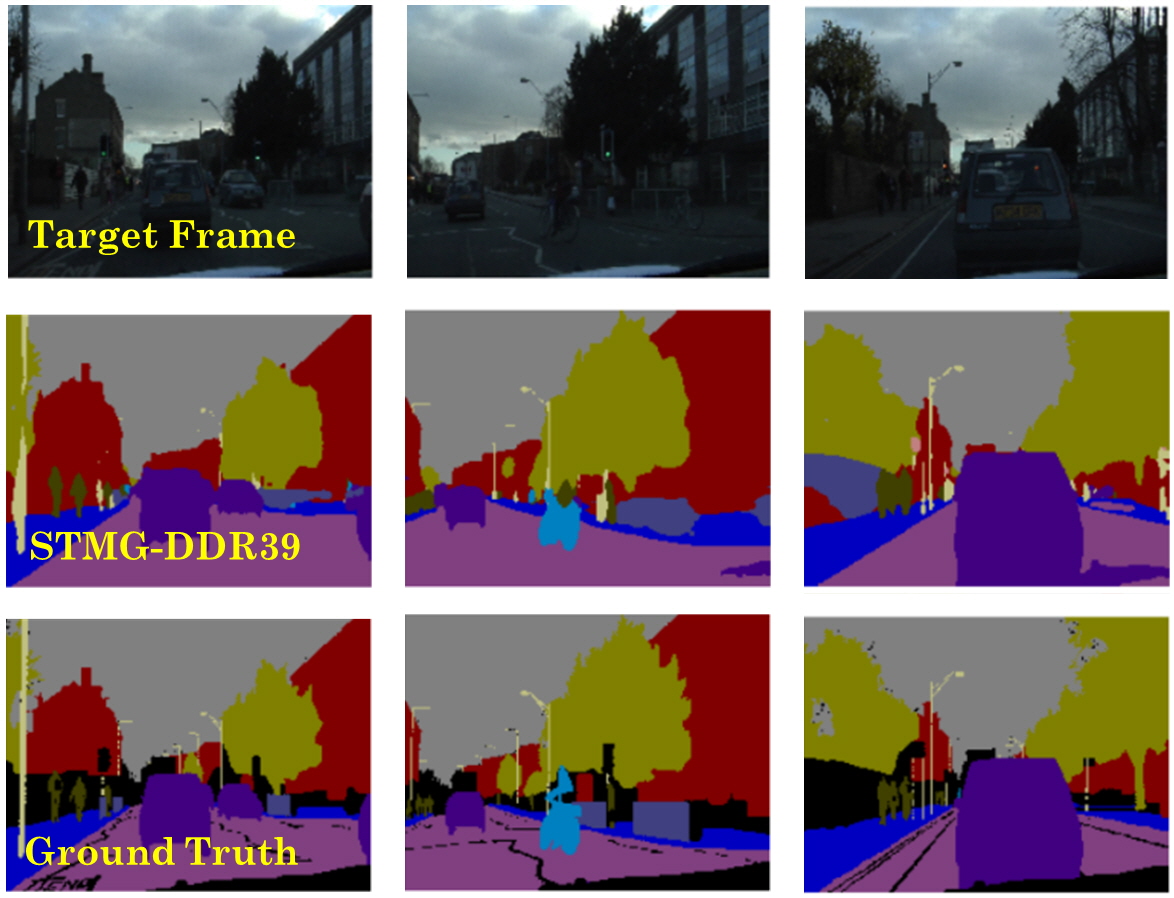}
        \vspace{-0.2in}
        \caption{\small \textbf{Qualitative results on CamVid.}}
        \label{fig:9}
        \vspace{-0.2in}
    \end{minipage}
    \hfill
    \begin{minipage}{0.35\linewidth}
        \centering
        \vspace{-0.12in}
        \captionof{table}{\small Performance comparison with real-time (> 30 FPS) semantic segmentation models on CamVid~\citep{brostow2009semantic} without pretraining on Cityscapes~\citep{cordts2016cityscapes}.}
        \vspace{-0.06in}
        \resizebox{1\textwidth}{!}{
        \renewcommand{\arraystretch}{1.05}
        \renewcommand{\tabcolsep}{1.0mm}
        \begin{tabular}{lcccc}
        \toprule
        Model & mIoU & FPS & Resolution \\
        \midrule
        BiSeNetV2~\citep{yu2021bisenet} & 72.4 & 124.5 & 720x960 \\
        GAS~\citep{lin2020graph} & 72.8 & 153.1 & 720x960 \\
        PP-LiteSeg-T~\citep{peng2022pp} & 73.3 & \bf 222.3 & 720x960 \\
        SFNet-RN-18~\citep{li2020semantic} & 73.8 & 35.5 & 720x960 \\
        STDC2-Seg~\citep{fan2021rethinking} & 73.9 & 152.2 & 720x960 \\
        MSFNet~\citep{si2019real} & 75.4 & 91.0 & 768x1024 \\
        HyperSeg~\citep{nirkin2021hyperseg} & \bf 78.4 & 38.0 & 576x768 \\
        \midrule
        DDRNet-39~\citep{hong2021deep} & 76.7 & 59.1 & 720x960 \\
        \textbf{- STMG-DDR39} & 75.5 & 70.2 & 720x960 \\
        \bottomrule
        \end{tabular}}
        \label{tab:camvid}
        \vspace{-0.2in}
    \end{minipage}
    \hfill
    \begin{minipage}{0.22\linewidth}
        \centering
        \vspace{-0.11in}
        \captionof{table}{\small Performance comparison with video semantic segmentation models on CamVid~\citep{brostow2009semantic}.}
        \vspace{-0.06in}
        \resizebox{1\textwidth}{!}{
        \renewcommand{\arraystretch}{1.05}
        \renewcommand{\tabcolsep}{1.0mm}
        \begin{tabular}{lcccc}
        \toprule
        Model & mIoU & FPS \\
        \midrule
        DFF~\citep{zhu2017deep} & 66.0 & 9.8 \\
        GRFP(5)~\citep{nilsson2018semantic} & 66.1 & 4.3 \\
        ACCEL-18~\citep{jain2019accel} & 66.7 & 5.9 \\    
        NetWarp~\citep{gadde2017semantic} & 67.1 & 2.8  \\
        ACCEL-50~\citep{jain2019accel} & 67.7 & 4.2 \\
        FANet-18~\citep{hu2020real} & 69.0 & \bf 154 \\
        FANet-34~\citep{hu2020real} & 70.1 & 121 \\
        TD4-PSP18~\citep{hu2020temporally} & 72.6 & 25.0 \\ 
        TD2-PSP50~\citep{hu2020temporally} & 76.0 & 11.1 \\
        \midrule
        \textbf{STMG-DDR39} & 75.5 & 70.2 \\
        \bottomrule
        \end{tabular}}
        \label{tab:camvid_video}
        \vspace{-0.2in}
    \end{minipage}
\end{figure}
\vspace{+0.05in}

\paragraph{Spatial Masking on CamVid}

We first learn the spatial masks using knowledge distillation as described in the main paper. However, we found that it was difficult to learn spatial masks since the amount of distortion between adjacent frames is too large in CamVid due to the low frame rate. Thus, we utilize a fixed spatial mask value to aggregate the features from previous and current frames. Specifically, the feature aggregation is computed as follows:
\begin{equation}
    \pi^{\prime}({{\bf I}_{cur}}) = m_{sp} \cdot \pi({{\bf I}_{cur}}) + (1 - m_{sp}) \cdot \pi({{\bf I}_{pre}}),
\end{equation}
where, $m_{sp}$ is fixed to $0.8$. We found that setting a higher $m_{sp}$ is crucial to compensate for information loss due to the low frame rate by using more features from current frame than in the previous frame.

\paragraph{Implementation Detail}

We set the initial learning rate to 0.001 and train the model for 968 epochs for DDRNet-39 and corresponding STMG module. We use a batch size of 4 with a cropping resolution of 720 × 960 for CamVid and conduct all experiments using PyTorch. For a fair comparison, we apply the same training setting to the corresponding STMG module. We train each model with two GeForce RTX 2080 Ti GPUs for DDRNet-39 and corresponding STMG module in the same setting.

\paragraph{Experimental Results}

We further report the experimental results on CamVid~\citep{brostow2009semantic} in Table~\ref{tab:camvid}, using DDRNet-39 backbone network~\citep{hong2021deep}. The results show that our method can achieve 70.2 FPS, achieving 18.8\% gain over the base model. We increase the FPS with less than 2\% drop in accuracy. That is, STMG provides a good trade-off comparable to state-of-the-art models on CamVid.

\section{Code and Dataset\label{sup/code}}

Code and dataset are available at \url{https://anonymous.4open.science/r/neurips2022stmg}.

\clearpage
\section{Additional Results on Cityscapes \label{sup/cityscapes}}
\vspace{-0.1in}

\begin{figure}[h]
    \centering        
    \includegraphics[width=0.95 \textwidth]{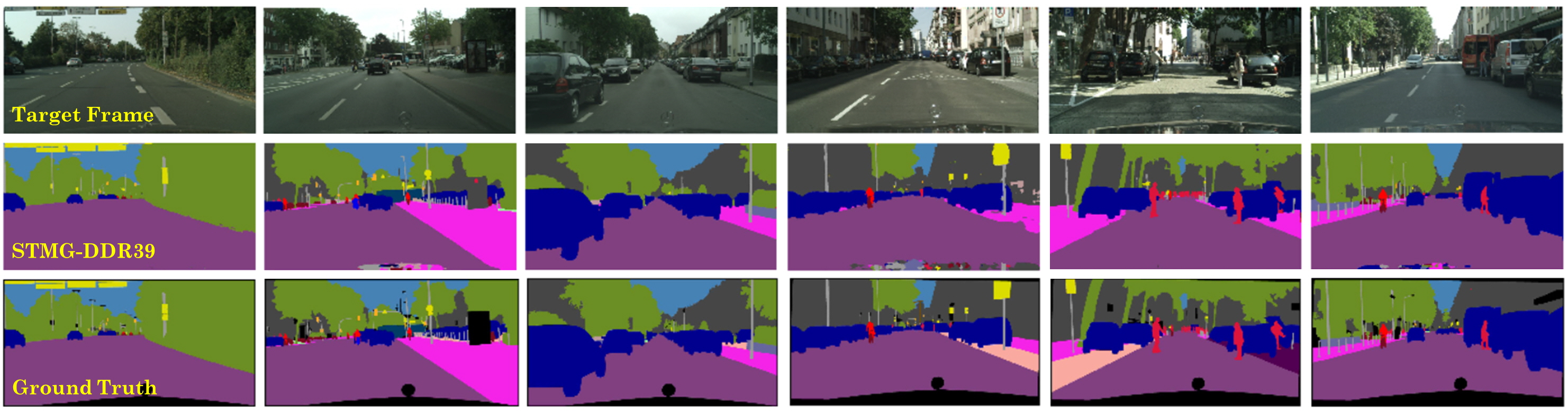}
    \caption{\textbf{Qualitative results on Cityscapes.} Visualization examples of our STMG. First row: Target Frame. Second row: STMG-DDR39. Third row: Ground Truth. Zoom in for best view.}
    \label{fig:10}
\end{figure}
\vspace{-0.1in}

\begin{figure}[h]
    \centering        
    \includegraphics[width=0.95 \textwidth]{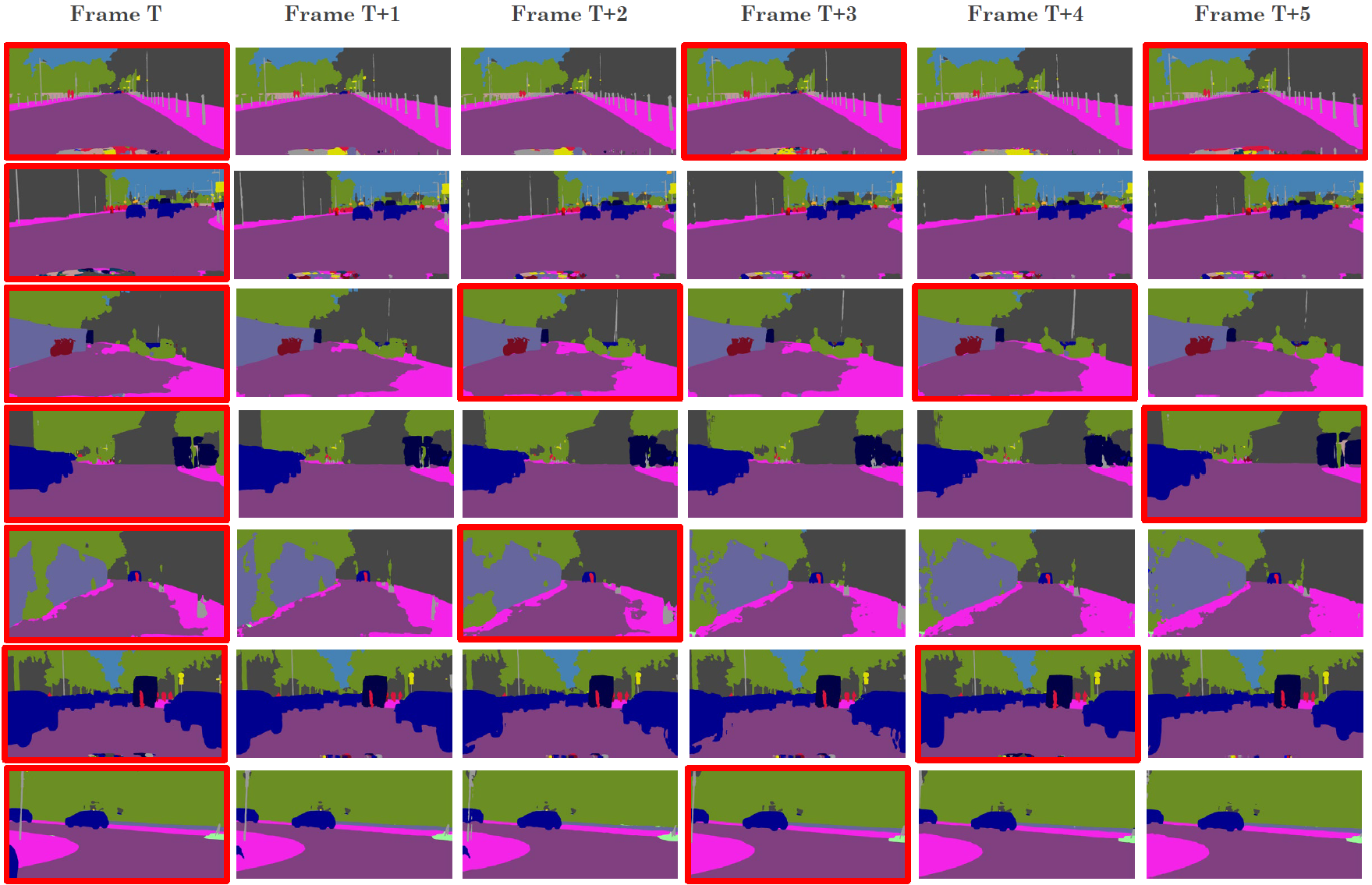}
    \caption{\textbf{Qualitative results of distortion-aware scheduling policy.} Frames highlighted in red represent key frames and other frames represent non-key frames. The fourth row shows that for static video streams, the scheduling policy continues to perform partial computations on non-key frames.}
    \label{fig:11}
    \vspace{+0.15in}
    
    \includegraphics[width=0.94 \textwidth]{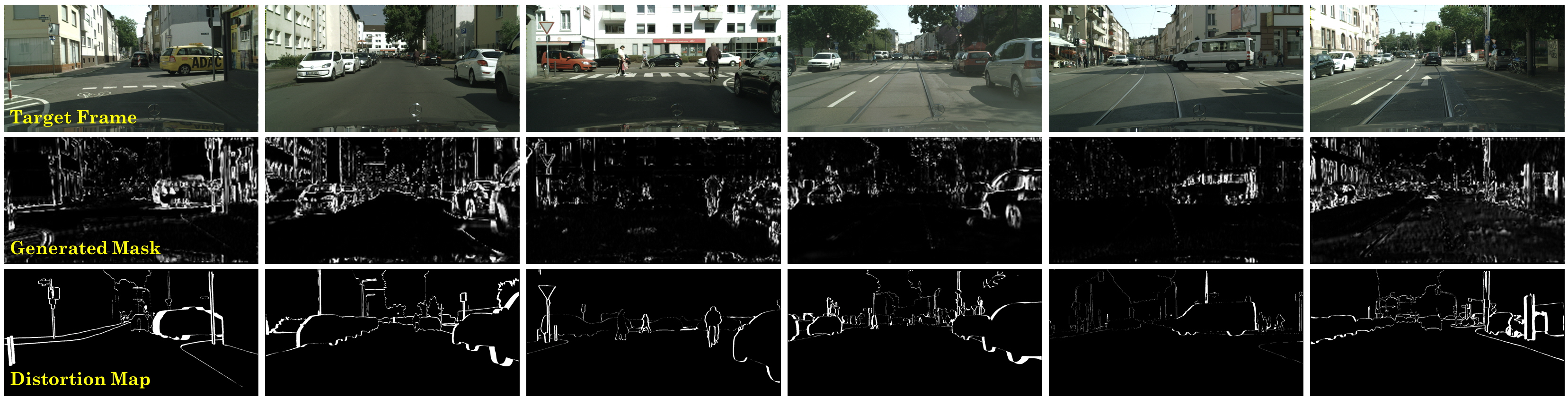}
    \vspace{+0.024in}
    \caption{\textbf{Qualitative results of generated spatial mask visualization on Cityscapes.} First row: Target Frame. Second row: Generated Mask. Third row: Distortion Map. Zoom in for best view.}
    \label{fig:12}
\end{figure}

\clearpage
\section{Additional Results on CamVid\label{sup/camvidadd}}

\vspace{-0.1in}
\begin{figure}[h]
    \centering        
    \includegraphics[width=0.95\textwidth]{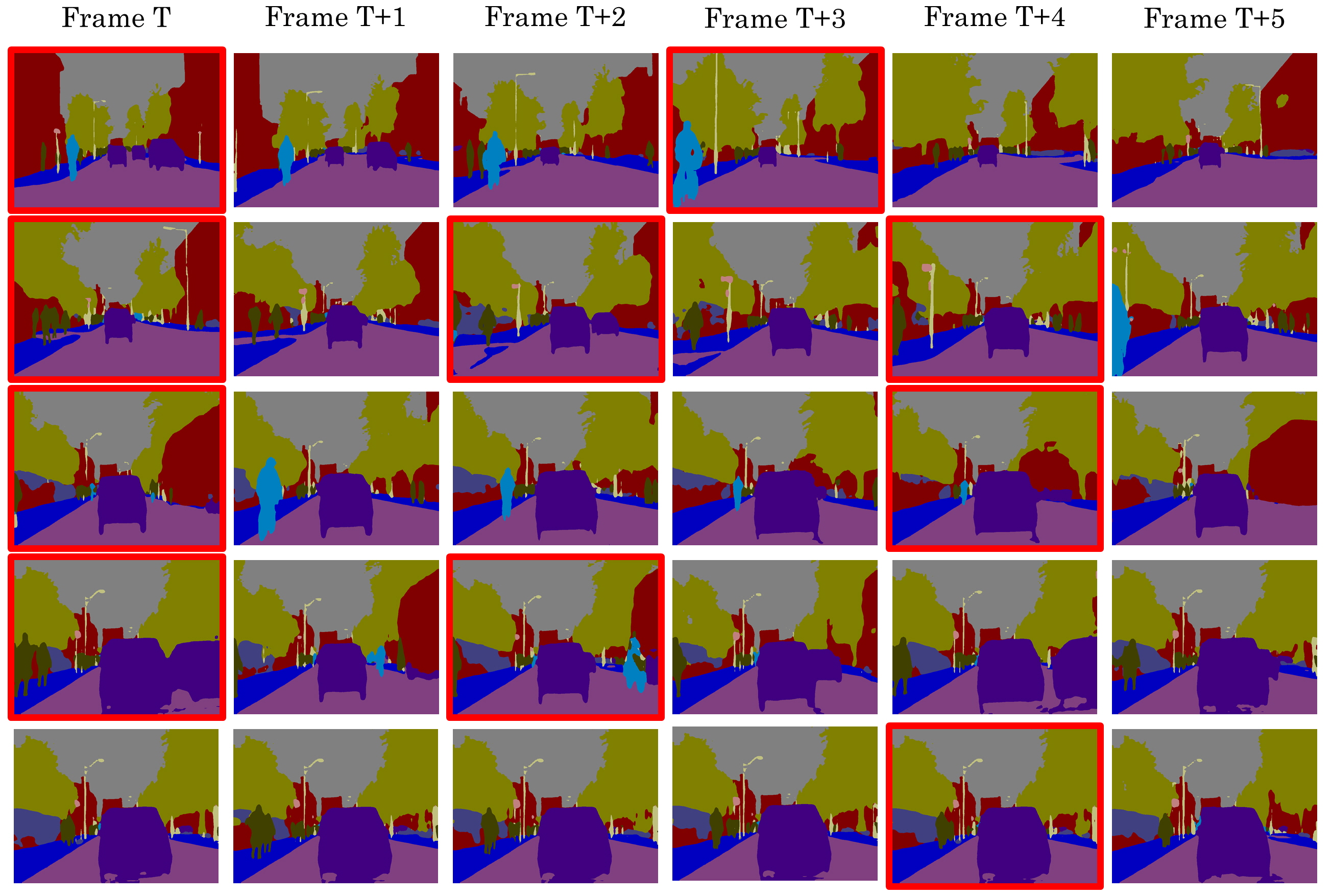}
    \vspace{-0.03in}
    \caption{\textbf{Qualitative results of distortion-aware scheduling policy on CamVid.} Frames highlighted in red represent key frames and other frames represent non-key frames. The fifth row shows that the scheduling policy continues to perform partial computations for static video streams.}
    \label{fig:13}
\end{figure}

\section{Discussion\label{sup/discussion}}

In this section, we discuss the limitation and potential negative societal impacts of our work.

\paragraph{Potential Societal Impacts} A potential ethical threat is that our proposed method could be exploited with malicious purpose to harm a society such as terrorism with limited computing resources (e.g. drones for terrorism). Further, our method could be exploited in mobile applications that can generate antisocial content, including real-time portrait matting for unethical deepfake videos. We hope that our proposed method would not be utilized for such malicious purposes to harm the society.

\paragraph{Limitation} In this work, we propose a novel framework for real-time speed up of video semantic segmentation models by exploiting the spatial-temporal locality of video. However, since our method targets real-time speed up and state-of-the-art real-time segmentation models are much faster than optical flow estimation, our method does not utilize flow-based feature propagation. Due to these properties, our method adopts a scheduling policy with a relatively short duration between key and non-key frames compared to previous video segmentation frameworks, as errors propagate when non-key frames are repeated. Further, since DBB~\citep{lee2018adaptive}-inspired pruning method optimizes the segmentation network while learning the pruning mask, our method uses different weights for key and non-key frames. This results in memory inefficiency both in inference time and training time, as two networks have to be loaded on the GPU. Because training a segmentation network requires a large memory, training is performed only on two adjacent frames: a fixed key frame for the previous frame and a trainable non-key frame for the current frame. We leave this limitation as future work.

\end{document}